\documentstyle[fullname,a4,11pt]{article}
\title{``I don't believe in word senses''\thanks{Sue Atkins ---Past
President, European Association for Lexicography; 
General Editor, Collins-Robert English/French Dictionary;
Lexicographical Adviser, Oxford University Press---
responding to a discussion which assumed discrete and disjoint word
senses, at `The Future of the Dictionary' workshop,
Uriage-les-Bains, October 1994.}}

\author{Adam Kilgarriff\\
ITRI\\
University of Brighton\\
Lewes Road\\
Brighton BN2 4GJ}

\begin{document}
\maketitle

\begin{abstract}
Word sense disambiguation assumes word senses.  Within the
lexicography and linguistics literature, they are known to be
very slippery entities.  The paper looks at problems
with existing accounts of `word sense' and describes the
various kinds of ways in which a word's meaning can deviate from its core
meaning.  An analysis is presented in which word senses are
abstractions from clusters of corpus citations, in accordance with
current lexicographic practice.  The corpus citations, not the word
senses, are the basic objects in the ontology.  The corpus citations
will be clustered into senses according to the purposes
of whoever or whatever does the clustering.  In the absence of such
purposes, word senses do not exist.

Word sense disambiguation also needs a set of word senses to
disambiguate between.  In most recent work, the set has been taken
from a general-purpose lexical resource, with the assumption that the
lexical resource describes the word senses of English/French/\ldots,
between which NLP applications will need to disambiguate.  The
implication of the paper is, by contrast, that word
senses exist only relative to a task.  
\end{abstract}

\section{Introduction}

There is now a substantial literature on the problem of word
sense disambiguation (WSD).  The goal of WSD research is generally
taken to be  disambiguation between the senses given
in a dictionary, thesaurus or similar.  The idea is simple
enough and could be stated as follows: 
\begin{quote}
 Many words have more than one meaning.  When a person understands a
sentence with an ambiguous word in it, that understanding is built on the
basis of just one of the meanings.  So, as some part of the human
language understanding process, the appropriate meaning has been chosen
from the range of possibilities.
\end{quote}
\noindent Stated in this way, it would seem
that WSD might be a well-defined task, undertaken by a particular
module within the human language processor.  This module could then be
modelled computationally in a WSD program, and this program,
performing, as it did, one of the essential functions of the human
language processor, would stand alongside a
parser as a crucial component of a broad range of NLP applications.  This
point of view is clearly represented in \nocite{Cottrell:89} Cottrell (1989):
\begin{quote}
[Lexical ambiguity] is perhaps the most important problem facing an
NLU system.  Given that the goal of NLU is understanding, correctly
determining the meanings of the words used is fundamental. \ldots The
tack taken here is that it is important to understand how people
resolve the ambiguity problem, since whatever their approach, it
appears to work rather well. (p 1) 
\end{quote}

Word meaning is of course a venerable philosophical topic, and
questions of the relation between the signifier and the signified will
never be far from the theme of the paper.  However, philosophical
discussions have not addressed the fact of lexicography and the
theoretical issues raised by sense distinctions as marked in
dictionaries.  We often have strong intuitions about words having
multiple meanings, and lexicography aims to capture them,
systematically and consistently.  The philosophy literature does not
provide a taxonomy of the processes underpinning the intuition, nor
does it 
analyse the relations between the word sense distinctions a dictionary
makes and the primary data of naturally-occurring language.  This is a
gap
that this paper aims to fill.


I show, first, that Cottrell's construal of word senses is at
odds with theoretical work on the lexicon (section~2); then,
that the various attempts to provide the concept `word sense' with
secure foundations over the last thirty years have all been
unsuccessful (section~3).  I then consider the lexicographers'
understanding of what they are doing when they make decisions about a
word's senses, and develop an alternative conception of the word
sense, in which it corresponds to a cluster of citations for a word
(section~4).  Citations are clustered together where they exhibit
similar patterning and meaning.  The various possible relations
between a word's meaning potential and its dictionary senses are
catalogued and illustrated with corpus evidence.

The implication for WSD is that there is no reason to expect a single
set of word senses to be appropriate for different NLP applications.
Different corpora, and different purposes, will lead to different
senses. 
In particular, the sets of word senses presented in different
dictionaries and thesauri have been prepared, for various purposes,
for various human users: there is no reason to believe those sets are
appropriate for any NLP application.

\section{Thesis and antithesis: practical WSD and theoretical
lexicology} 

\subsection{Thesis}

NLP has stumbled into word sense ambiguity.  

Within the overall shape of a natural language understanding system
--morph\-o\-logical analy\-sis, pars\-ing, seman\-tic and prag\-matic
inter\-pret\-ation-- word sense ambiguity first features as an irritation.
It does not appear as a matter of particular linguistic interest, and
can be avoided altogether simply by treating all words as having just
one meaning.  Rather, it is a snag: if you have both river {\em
bank} and money {\em bank} in your lexicon, when you see the word {\em
bank} in an input text you are at risk of selecting the wrong one.
There is a practical problem to be solved, and
since Margaret Masterman's group started examining it in the 1950s
(see, e.g., \nocite{Sparck-Jones:86} Sparck Jones (1986)), people have been writing
programs to solve it. 


NLP has not found it easy to give a very principled answer to the
question, ``what goes in the lexicon''.  Before the mid-1980s, many
systems made no claims to wide coverage and contained only as many
words in the lexicon as were needed for the `toy' texts that were
going to be analysed.  A word was only made ambiguous --- that is,
given multiple lexical entries --- if it was one that the researchers
had chosen as a subject for the disambiguation study.  This was
clearly not an approach that was sustainable for wide coverage systems, and
interest developed in dictionaries, as relatively principled,
wide-coverage sources of lexical information.

As machine-readable versions of dictionaries started to become
available, so it became possible to write experimental WSD programs on
the basis of the dictionary's verdict as to what a word's senses were
\cite{Lesk:86,Jensen:87,Slator:88,Veronis:90,Guthrie:90,Guthrie:91,Dolan:94}. Looked at the other
way round, WSD was one of the interesting things you might be able to
do with these exciting new resources.

Since then, with the advent of language corpora and the rapid growth
of statistical work in NLP, the number of
possibilities for how you might go about WSD has mushroomed, as has
the quantity of work on the subject
\cite{Brown:91,Hearst:91,McRoy:92,GCY:92,Yarowsky:92,GCY:93}.
\nocite{Clear:94} Clear (1994), \nocite{SchutzePederson:95} Sch\"{u}tze
and Pederson (1995) and
\nocite{Yarowsky:95} Yarowsky (1995) are of particular interest
because of 
their
approach to the issue of the set of word senses to be disambiguated
between.   Sch\"{u}tze and Pederson devised high-dimensionality
vectors to describe the context of each occurrence of their target
word, and then clustered these vectors.  They claim that the
better-defined of these clusters
correspond to word senses, so a new occurrence of the word can be
disambiguated by representing its context as a vector and identifying
which cluster centroid the vector is  closest to.  This system has the
characteristic that a context may be close to more than one cluster
centroid, so at times it may be appropriate to classify it as more
than one sense.

Both \nocite{Clear:94} Clear (1994) and \nocite{Yarowsky:95} Yarowsky
(1995) provide a
mechanism for the user to input the senses between which they would
like the system to disambiguate.  They ask the user to classify a
small number of statistically-selected `seed' collocates, so the user
determines the senses to be disambiguated between when deciding on the
senses he or she will assign seed collocates to.\footnote{In
Yarowsky's work, this is just one of the options for providing seeds
for the process.}  Clear then finds all the words which tend to
co-occur with the node word in a large corpus, and quantifies, for a
very large number of words, the evidence that it occurs with each of
the seeds, and thus indirectly, with each sense of the nodeword.
Disambiguation then proceeds by summing the evidence for each sense
provided by each context word.  

Yarowsky's method is iterative: first,
those corpus lines for the nodeword which contain one of the seed
collocates are classified.  Then the set of corpus lines so classified
is examined for further indicators of one or other of the senses of
the word.  These indicators are sorted, according to the strength of
evidence they provide for a sense.  It will now be possible to
classify a larger set of corpus lines, so producing more indicators
for each sense, and the process can be continued until all, or an
above-threshold proportion, of the corpus lines for the word are
classified.  The ordered list of sense-indicators will then serve as a
disambiguator for new corpus lines.

In the Semantic Concordance project at Princeton a lexicographic team
has been assigning a WordNet \cite{Miller:90} sense to each noun,
verb, adjective and adverb in a number of texts, thus providing a
`gold standard' disambiguated corpus which can be used for training
and evaluating WSD programs \cite{Landes:96}.

In 1994-95, there was an extended discussion of whether WSD
should be one of the tasks in the MUC program.\footnote{The MUC (Message
Understanding Conference) is a series of US
Government-funded, competitive, quantitatively-evaluated exercises in
information extraction \cite{MUC5}.} This would have provided for
competitive evaluation of different NLP groups' success at the WSD
task, as measured against a `benchmark' corpus, in which each word had
been manually tagged with the appropriate WordNet sense number (as in
the Semantic Concordance).  
Some trials took place, but the decision was not to
proceed with the WSD task as part of the 1996 MUC6 evaluation, as
there was insufficient time to debate and define detailed policies.
The theme has recently been taken up by the Lexicons Special Interest Group
of the Association for Computational Linguistics, and a pilot
evaluation exercise is taking place in 1998: a milestone on
the road from research to technology.

\subsection{Antithesis}

Since the publication of {\em Metaphors We Live By}
\cite{Lakoff:80} and {\em Women, Fire and Dangerous Things}
\cite{Lakoff:87}, there has been one approach to linguistics --
cognitive linguistics -- for which metaphor has been a
central phenomenon.  Metaphor is, amongst other things, a process
whereby words spawn additional meanings, and cognitive linguists
are correspondingly
interested in polysemy. Lakoff's analysis of the polysemy of {\em
mother} is hugely cited.  Word sense ambiguity can often be
seen as a trace of the fundamental processes underlying language
understanding \cite{Sweetser:90}.  The structures underlying the
distinct meanings of words are at the heart of the cognitive
linguistics enterprise \cite{Geeraerts:90,JTaylor:89}.

Working in this framework, \nocite{Cruse:95} Cruse (1995) gives a detailed
typology of polysemy.  He distinguishes polysemy, defined according to
distinctness of meaning, from polylexy, which is where, in addition to
distinctness of meaning, distinct lexical entries are required.  A word
is polysemous but not polylexic where its non-base meanings are
predictable, so they can be generated as required and need not be stored.
He also addresses where readings are antagonistic and where they are
not, and the characteristics of the different semantic properties, or
`facets', of a sense.  He uses ambiguity tests to tease out a number
of issues, and a full Cruse lexical entry would contain: a
specification of polysemous senses; their lexical relations including
their relations to each other; whether they were antagnostic or not;
the facets, shared or otherwise, of each, and the extent to which
distinct facets of meaning could operate autonomously, so approach the
status of senses on their own.  He considers several varieties of
`semi-distinct' readings.

Lexical ambiguity has also moved centre-stage within theoretical and
computational 
linguistics.  Both AAAI and ACL
have recently devoted workshops to the topic.\footnote{The AAAI Spring
		  Symposium on Representation and Acquisition of
		  Lexical Information, Stanford, April 1995 and the
		  ACL SIGLEX Workshop on The Breadth and Depth of
		  Semantic Lexicons, Santa Cruz, June 1996.} When Pustejovsky and others discuss the generative lexicon
\cite{Pustejovsky:91,Briscoe:90}, the generative
processes they have in mind are, again, ones whereby words spawn
additional meanings (or, at least, additional uses).  Regular polysemy
\cite{Apresjan:74} has recently been discussed, and computational
mechanisms for addressing it proposed, by \nocite{Ostler:91} Ostler
and Atkins (1991),
\nocite{Lyons:95} Lyons (1995) and \nocite{Copestake:95} Copestake and
Briscoe (1995), {\em inter alia}.
Levin and colleagues have also been finding systematicity in lexical ambiguity, in
relation to verb classes, their patterns of subcategorisation, and
their patterns of alternation
\cite{LR:91,Levin:93,LevinSongAtkins:97}.

This combination of circumstances leads to an odd situation.  Much WSD
work proceeds on the basis of there being a computationally relevant,
or useful, or interesting, set of word senses in the language,
approximating to those stated in a dictionary. To the WSD community,
word senses are, more or less, as the dictionary
says.\footnote{Sometimes not all the sense distinctions recognised in
the dictionary are viewed as salient to the program.  WSD researchers
tend to be lumpers, not splitters \cite{Dolan:94}.}  (This is not, of
course, to say that WSD authors have not noted the theoretical
problems associated with dictionary's word senses.)  WSD research has
gone a long way on this basis: it is now common for papers to present
quantitative comparisons between the performance of different systems.
Meanwhile, the theoreticians provide various kinds of reason to
believe there is no such set of senses. To get beyond this impasse, we
need to look more closely at the question, ``what is a word sense?''.

\section{What is a word sense?}
\begin{quote}
No entity without identity. \cite{Quine:69}
\end{quote}
\noindent Or, to know what something is, is to know when something is it.  To
know what a word sense $s_1$ is, is to know which uses of the word are part of $s_1$ and which are not, probably because they are
part of $s_i$ where $i\neq1$.  If we are to know what word senses
are, we need operational criteria for distinguishing them. 

\subsection{Selection and modulation}

A good starting point is Cruse's textbook on Lexical Semantics
\cite{Cruse:86}.  `Lexical units' are the object of his enquiry, and
he devotes two substantial chapters to specifying what they are.  He 
states the heart of the problem thus:

\begin{quote}
One of the basic problems of lexical semantics is the apparent
multiplicity of semantic uses of a single word form (without
grammatical difference).
\end{quote}
He addresses in some detail
the difference between those cases where the context {\bf selects} a
distinct unit of sense, from those where it {\bf modulates}
the meaning.  In the pair
\begin{quote}
Have you put the money in the bank?\\
The rabbit climbed up the bank.
\end{quote} 
\noindent the two sentences select different meanings of {\em bank}, whereas in
\begin{quote}
He doesn't often oil his bike.\\
Madeleine dried off her bike.\\
Boris's bike goes like the wind.
\end{quote} 
\noindent different aspects of the bicycle ---its mechanical
parts; its frame, saddle and other large surfaces; its (and its
rider's) motion--- are
highlighted in each case.  The meaning of {\em bike} is modulated
differently by each context.\footnote{Cruse identifies two major
varieties of modulation, of which highlighting is one.}

\subsection{Ambiguity tests}

The selection/modulation distinction is closely related to the
distinction between ambiguity and generality, also referred to as
`vagueness', `indeterminacy' and `lack of specification'.\footnote{See
\nocite{Zwicky:75} Zwicky and Sadock (1975) for a fuller discussion of the terms and their
sources.}  Where a word is ambiguous, a sense is selected.  Where a
word-meaning is general between two readings, any particular context
may or may not modulate the word-meaning to specify one or other of
the readings. Thus, {\em hand} is unspecified between right hands and
left hands; some sentences modulate the meaning to specify a right or
left hand, as in ``When saluting, the hand should just touch the
forehead'', while others do not.\footnote{Also related to this
distinction is the polysemy/homonymy distinction: when do we have two
distinct words, and when, one word with two meanings?  Most
commentators agree that there is a gradation between the two, with the
distinction being of limited theoretical interest.  For some purposes,
the distinction may be more useful than the vagueness/ambiguity one
\cite{Krovetz:96}.  In practice, similar difficulties arise in
distinguishing homonymy from polysemy, as in distinguishing vagueness
from ambiguity.}

Clearly, {\em bank} is ambiguous between the readings demonstrated
above; {\em bike} is not.  But for many reading-pairs, the answer is not
clear:\footnote{
The examples are taken by comparing four state-of-the-art
English learners' dictionaries \cite{LDOCE3,OALDCE5,COBUILD2,CIDE}
and finding words where the lexicographers in one team made one
decision regarding what the distinct word senses were, whereas those in
another made another.  This immediately has the effect of introducing
various factors which have not been considered in earlier theoretical
discussions.}

\begin{itemize}
\item I planted out three rows of beans yesterday.\\
Cook the beans in salted water. 
\item The cottage was charming. \\ Our hosts were charming.
\item Bother! I was about to talk to John, but now he's disappeared!
({\sc not-here})\\
I can't find it anywhere, it seems to have disappeared. ({\sc can't-find})
\end{itemize}

A number of tests have been proposed for determining whether a word is
ambiguous or general between two meanings.  They are catalogued in
\nocite{Zwicky:75} Zwicky and Sadock (1975), 
\nocite{Cruse:86} Cruse (1986),
and \nocite{Geeraerts:93} Geeraerts (1993).  Here, I shall describe
only one of the 
more successful tests, the `crossed readings' one.

\begin{quote}
Mary arrived with a pike and so did Agnes.
\end{quote}
\noindent could mean that each arrived with a carnivorous fish, or that each
arrived bearing a long-handled medieval weapon, but not that the  one
arrived with 
the fish and the other with the weapon.  On the other hand, in
\begin{quote}
Tom raised his hand and so did Dick.
\end{quote}
\noindent each might have raised a right hand, each might have raised
a left, or one might have raised his right, and the other, his left.  The
question now is, in
\begin{quote}
Ellen bought some beans, and so did Harry.
\end{quote}
\noindent is it possible that Ellen bought plants and Harry,
food?
If so, then the conclusion to be drawn from the test is that {\em bean} is
ambiguous between the readings, and if not, then it is
not.\footnote{For many putatively ambiguous reading-pairs, there are 
intermediate cases.  A sprouting bean, or one bought for planting, is
intermediate between {\sc food} and {\sc plant}.  But the possibility
of intermediate cases does not preclude ambiguity: whether two
readings of a word are completely disjoint, permitting no intermediate
cases, is a different question to whether a word is ambiguous.  This
imposes a further constraint on ambiguity tests.  A speaker might say,
``Ellen and Harry must have bought the same kind of {\em bean},
unless, say, Ellen bought plants and Harry bought beans sold at the
supermarket but which he was intending to plant''.  We should not
infer that {\em bean} is vague.  Rather, we must insist that both of
the crossed readings are prototypical.  (There are of course further
difficulties in making this constraint precise).}

\subsubsection{Criticisms of the tests}

The tests are generally presented with the aid of an unproblematical example
of ambiguity and an unproblematical example of vagueness. This is done in
order to demonstrate what the test is and what the two contrasting
outcomes are. However, this is not to use the tests in anger.
What we want of a test is that it is consistent with our intuitions, where our
intuitions are clear, and that it resolves the question, where our intuitions
are unclear. The cross-reading
test fares tolerably well in meeting the consistency
condition (though see \cite{Geeraerts:93} for a contrary view). But do
the tests help where intuitions are unclear? There is little if any
evidence that they do. Here I discuss three classes of problems.

Firstly, it must be possible to construct a plausible
test sentence.  The word in its two uses must be able to occur with
the same syntax and the same lexico-grammatical environment.  Consider the
transitive and intransitive uses of {\em eat}, as in ``John ate the apple''
and ``John ate''. Is this a case of ambiguity or vagueness?
\begin{quote}
*Mary ate, and John, the apple.
\end{quote}
\noindent is unacceptable, but the reason is that elided constituents must have the same
syntax and subcategorisation in both their expressed and elided occurrences.
It might be desirable to treat all words with alternative subcategorisation
possibilities as ambiguous. But whether or not that is
done, the test still fails to elucidate on the topic of a word's meaning,
where the word has different syntax in different uses. The test can only be
posed where the two uses are syntactically similar.  

The {\em
disappear} example displays a different variant of this problem.  The
{\sc can't-find} and {\sc not-here} readings have different aspectual
characteristics: {\sc
can't-find} is stative while {\sc not-here} is a punctual
`achievement' verb.  
\begin{quote}
Martha disappeared and so did Maud.
\end{quote}
\noindent does not permit a crossed reading, but that is because we cannot
construct a viable aspectual interpretation for the conjoined sentence,
compare
\begin{quote}
? I evicted and knew her.\footnote{Eight out of ten informants found
the related sentence, ``I loved and married her'', odd.  The two who
found it acceptable were reading {\em and} as an indicator of
temporal sequence.}
\end{quote}
\noindent It is not evident whether there is a conclusion to
be drawn regarding polysemy.

In general, one can apply more or less effort into trying to find a
test sentence (and associated context) in which the crossed reading is
plausible.  A test is clearly flawed, if, the more ingenuity the 
investigator displays, the more of one particular outcome he or she
will get.  
(The crossed reading test is the test which suffers least
from this flaw, but it is nonetheless in evidence.)

The second point is more general and theoretical. A certain amount of
interpretation of an utterance must have been undertaken before an
acceptability judgement can be made. Three parts of the interpretation
process are lexical access, parsing, and `pragmatic interpretation',
the final stage of incorporating the new information into the
discourse model.  The premise behind acceptability judgements is that
a subject can report on the outcome of the first two stages,
irrespective of what goes on in the third.  For a wide range of
syntactic questions, the methodology is widely used and has proved its
worth.

\nocite{Nunberg:78} Nunberg's (1978) arguments illustrate the hazards of the premise
for questions in lexical semantics.  Consider
\begin{quote}
The newspaper costs 25p and sacked all its staff.
\end{quote}
\noindent It is anomalous. We cannot place the origin of the anomaly in the
lexicon unless we grant the word two lexical entries, one for a copy
of the newspaper and one for the owner or corporate entity. Then the
size of our lexicon will start to expand, as we list more and more of
the possible kinds of referent for the word, and still it will never
be complete. So the origin of the anomaly must be the interpretation
process.  But the anomaly seems similar to the anomaly that occurs
with {\em bank}. In a case lying between {\em newspaper} and {\em
bank}, how would we know whether the source of the anomaly was the
lexicon or the interpretation process?  In the general case the point
at which the lexical process becomes a general-purpose interpretative
one cannot be identified. There is no accessible intermediate
representation in which lexical ambiguities are resolved (for
acceptable sentences) but in which the contents of the sentence has
not been incorporated into the hearer's interpretation of the
discourse.  \nocite{Geeraerts:93} Geeraerts (1993) presents an extensive critique of
the tests along these lines, presenting evidence that the different
tests give contradictory results, and that even if we constrain
ourselves to looking at just one of the tests, they can all be made to
give contradictory results by manipulating the context in which the
item under scrutiny is set.

The third problem is simply the lack of evidence that the tests give
stable results.  It will sometimes happen that, for the same
reading-pair, an informant will deem crossed readings possible for
some test sentences and not for others.  Or different informants will
have conflicting opinions.  There are, remarkably, no careful
discussions of these issues in the literature.  The merit of the
method of acceptability judgements for syntax rests on the relative
stability of their outcomes: they work (to the extent they do) because
linguists agree where the stars belong.  Preliminary investigations
into the stability of outcomes in lexical semantics suggest that it is
severely lacking.

\subsection{Psycholinguistics and `semantic priming'}

There is a set of findings in psycholinguistics which might allow us
to base an account of `word sense' directly on the mental lexicon.
The experimental paradigm is called `semantic priming'.  It is
well-established that, if I have just heard the word {\em doctor} (the
`prime'), and then a sequence of letters (the `target') is flashed up
on a screen and I am asked to identify whether it is a word or not, I
respond faster if it is a word and it is {\em nurse} than if it
is a word but unrelated to {\em doctor}.\footnote{This is the `lexical
decision' task in a mixed, visual and auditory procedure. It is one of
a variety of versions of semantic priming experiments.  The basic
effect is robust across a number of experimental strategies.}  

If an ambiguous
prime such as {\em bank} is given, it turns out that both {\em river}
and {\em money} are primed for.  If {\em bank} is presented in isolation, priming for both {\em river}
and {\em money} is found for another second or two.  In a context which serves to make only one of these appropriate, after something between
50 and 200 ms a choice is made and after that only the appropriate target
is primed for.   

So, for ambiguous words, priming behaviour has a distinct `signature'.
Perhaps it is possible to identify whether a word is vague or
ambiguous by seeing whether it exhibits this signature.

The hypothesis is explored by \nocite{Williams:92} Williams (1992).  He looked at
adjectives, for example {\em firm}, for which the two readings were
represented by {\em solid} and {\em strict}.  After confirming that
the prime, {\em firm}, in isolation, primed equally for  {\em solid} and {\em
strict}, he tested to see
if {\em solid} was primed for when {\em firm} occurred in a
{\sc strict} context, and {\em vice versa}, after delays of 250, 500 and
850 ms.   

His results were asymmetrical.  He identified central meanings ({\sc
solid}) and non-central ones ({\sc strict}).  Where the context
favoured the central reading, the non-central-sense targets were not
primed for.  But when the context favoured the non-central reading,
central targets were.  The experiments provide evidence that the
various meanings of polysemous words are not functionally independent
in language comprehension, and that not all senses are equal, in their
representation in the mental lexicon.  Williams discusses the
asymmetrical results in terms of hierarchical meaning structures.

Priming experiments do show potential for providing
a theoretical grounding for distinguishing ambiguity and generality, but
more work needs to be done, and the outcome would not be a simple, two-way,
ambiguous/general distinction.  Also, the method would never be
practical for determining the numbers of senses for a substantial
number of words.  The results of the experiments are just not
sufficiently stable: as Williams says, the priming task ``suffers
from a large degree of item and subject variability'' (p 202).

\section{Lexicographers, dictionaries, and authority}

\begin{quote}
What set of procedures do lexicographers have available to them to pin
down those protean entities, `meanings'?  Faced with the almost
unimaginable diversity of the language they are trying to describe,
with the knowledge that what for the sake of convenience we are
pleased to call a language is in many ways a synthesis of shifting
patterns that change from year to year, from locality to locality,
from idiolect to idiolect, how do they arrive at those masterpieces of
consensus, dictionaries?  How do they decide what, for the purposes of
a dictionary, constitutes the meaning of a word, and where, in the
case of polysemous words, one meaning ends and the next begins?
\cite[p 89]{Ayto:83}
\end{quote}

In the middle of this debate stand the lexicographers.  The word
senses that most WSD researchers aim to discriminate
are the product of their intellectual 
labours.  But this is far from the purpose for which the dictionary
was written.

Firstly, any working lexicographer is well aware that, every day, they
are making decisions on whether to `lump' or `split' senses that are
inevitably subjective:\footnote{ {\bf Lumping} is considering two slightly
different patterns of usage as a single meaning. {\bf Splitting} is
the converse: dividing
or separating them into different meanings.}
frequently, the alternative decision would have been equally valid.
In fact, most dictionaries encode a variety of relations in the grey
area between ``same sense'' and ``different sense'': see
\nocite{ak:chum} Kilgarriff (1993) for a description of the seven methods used in
\cite{LDOCE2}.

Secondly, any particular dictionary is written with a particular
target audience in mind, and with a particular editorial philosophy in
relation to debates such as `lumping {\em vs.} splitting', so the notion
of specifying a set of word senses for a language in isolation from
any particular user group will be alien to them.

Thirdly, many are aware of the issues raised
by Lakoff, Levin, Pustejovsky and others, with several
lexicographers bringing valuable
experience of the difficulties of sense-division to that
literature (see below). 

Fourthly, the weight of history: publishers expect to publish,
bookshops expect to sell, and buyers expect to buy and use
dictionaries which, for each word, provide a (possibly nested) list of
possible meanings or uses.  Large sums of money are invested in
lexicographic projects, on the basis that a dictionary has the
potential to sell hundreds of thousands of copies.  Investors will not
lightly adopt policies which make their product radically different to
the one known to sell.  However inappropriate the nested list might be
as a representation of the facts about a word, for all but the most
adventurous lexicographic projects, nothing else is
possible.\footnote{The format of the dictionary has remained fairly
stable since Dr.\ Johnson's day.  The reasons for the format, and the
reasons it has proved so resistant to change and innovation, are
explored at length in \cite{Nunberg:94}. In short, the development of
printed discourse, particularly the new periodicals, in England in the
early part of the eighteenth century brought about a re-evaluation of
the nature of meaning.  No longer could it be assumed that a
disagreement or confusion about a word's meaning could be settled
face-to-face, and it seemed at the time that the new discourse would
only be secure if there was some mutually acceptable authority on what
words meant.  The resolution to the crisis came in the form of
Johnson's Dictionary.  Thus, from its inception, the modern dictionary
has had a crucial symbolic role: it represents a methodology for
resolving questions of meaning. Hence ``the dictionary'', with its
implications of unique reference and authority (cf.\ ``the Bible'')
\cite{Leech:81}.  Further evidence for this position is to be found in
\nocite{McArthur:87} McArthur (1987), for whom the ``religious or quasi-religious
tinge'' (p 38) to reference materials is an enduring theme in their
history; \nocite{Summers:88} Summers (1988), whose research into dictionary use found
that ``settl[ing] family arguments'' was one of its major uses (p 114,
cited in  \nocite{Bejoint:94} Bejoint (1994), p 151); and \nocite{Moon:89} Moon (1989) who
catalogues the use of the UAD (Unidentified Authorising Dictionary)
from newspapers letters pages to restaurant advertising materials (pp
60--64).

The implications for ambiguity are this: to solve disputes about
meaning, a dictionary must be, above all, clear.  It must draw a line
around a meaning, so that a use can be classified as on one side of
the line or the other.  A dictionary which dwells on marginal or vague
uses of a word, or which presents its meaning as context-dependent or
variable or flexible, will be of little use for purposes of settling
arguments.  The pressure from this quarter is for the dictionary to
present a set of discrete, non-overlapping meanings for a word, each
defined by the necessary and sufficient conditions for its application
---whatever the facts of the word's usage.}

The division of a word's meaning into senses is forced onto
lexicographers by the economic and cultural setting within which they
work.  Lexicographers are obliged to describe words as if all words
had a discrete, non-overlapping set of senses.  It does not follow
that they do, nor that lexicographers believe that they do.

\subsection{Lexicographical literature}

Lexicographers write dictionaries rather than writing about writing
dictionaries.  Little has been written that answers the challenge
posed by Ayto in the quotation above.
Zgusta's influential {\em Manual} (1971)\nocite{Zgusta:71}, while stating that the specification of word meaning is the
central task for the lexicographer (p 23) and
the division of a word's meanings into senses is a central part of that,
gives little guidance beyond admonishments to
avoid making too many, or too few, distinctions (pp 66--67).

Ayto's own offering in the 1983 paper is the `classical' or `analytic'
definition, comprising genus and differentiae. In choosing the genus term, the
lexicographer must take care to neither select one that is too general
\----{\em entity} would not do as a genus term for {\em tiger}\---- nor too
specific, if the specific genus term is likely to be unknown by the
dictionary users. Where two meanings of a word have different genus terms,
they need treating as different senses.
The next task is to identify the differentiae required to separate out senses
falling under the same genus term.  He
discusses {\em cup}, and argues that there are three senses, one for the
`trophy' sense, one for the varieties standardly made of china or earthenware,
and one for the prototypically plastic or paper varieties. But his
consideration of the arguments for treating the second and third of these as
distinct ends in a welter of open questions.

\nocite{Stock:83} Stock (1983) is a response to Ayto's piece, and finds it wanting, firstly,
in the circularity involved in using different genus terms to identify
distinct senses ---the lexicographer will only look for distinct genus terms
after determining there are distinct senses--- and secondly, in that the
model cannot be applied to many words.   She looks closely at
{\em culture}, noting how different dictionaries have divided the
territory that the word covers in quite different ways, and observes,
\begin{quote}
It is precisely the lack of clarity in our use of the word {\em
culture} which makes it such a handy word to have at one's disposal.
It offers, as it were, semantic extras just because in most uses its
possible meanings are not clearly disambiguated.  \ldots  What can 
the dictionary maker do to reflect this
state of affairs? \ldots They do not, cannot by their very structure,
show that there is slippage between some of the senses that they give
but not between others. (p.\ 139)
\end{quote}
\nocite{Hanks:94} Hanks (1994), looking at {\em climb}, and
\nocite{FillmoreAtkins:92} Fillmore and Atkins (1992),
studying the semantic field centred on {\em risk}, make similar
comments about the inadequacies of dictionary conventions,
and appeal to prototype theory and frame semantics for richer
frameworks to describe the relationships between the different ways a
word (or word-family) is used.

Stock, Hanks and Atkins were all involved in the early stages of the
COBUILD project, which, in the early 1980s, broke new ground in
lexicography through its use of very large computerised language
corpora
\cite{Sinclair:87}. Good lexicographic practice had long used huge
citation indexes, but being able to see hundreds of instances of a
word in context, ordinary and extraordinary examples thrown together,
was a radical development.  It has changed how lexicographers think
about meaning.  Where Ayto's paper offers semantic analysis, Stock
presents corpus evidence.   The lexicographer's primary source of
evidence for how a word behaves switches from subjective to 
objective; from introspection to looking at contexts.

\subsection{A corpus-based model of word senses}

This suggests a quite different answer to the question, ``what is a
word sense?''  Corpus lexicography proceeds approximately as follows.
For each word, the lexicographer

\begin{enumerate}
\item
calls up a concordance\footnote{By `concordance' I mean a display
which presents a line of context for each occurrence of the word
under scrutiny in the corpus, with all occurrences of the key word aligned.  Fuller
details are, of course, system specific, but it has rapidly become
evident that this
kind of display is the basic requirement for any corpus lexicography
system.}  for the word;
\item
divides the concordance lines into clusters, so that, as far as possible, all
members of each cluster have much in common with each other, and little in common with members of other clusters;
\item
for each cluster, works out what it is that makes its members belong
together, re-organising clusters as necessary;\item
takes these conclusions and codes them in the highly
constrained language of a dictionary definition.
\end{enumerate}

Putting the concordance lines into clusters is data-driven rather than
theory-driven. The lexicographer may or may not be explicitly aware of
the criteria according to which he or she is clustering.\footnote{The interactions between the lexicographers' clusters and the
automatic clusters produced for Information Retrieval purposes
\cite{SchutzePederson:95}, and the potential for automating some of the
clustering that the lexicographer performs, are subjects of current
research.}  (It is a requirement for corpus lexicography software
that it supports manual clustering \cite{Hector,xkwic}.)
Stage 3 is just a
fallible {\em post hoc} attempt to make the criteria explicit. The
senses that eventually appear in the dictionary are the result, at
several removes, of the basic
clustering process.  

Ambiguity tests failed to provide us with an account of what it meant
for two uses of a word to belong to the same word sense.  Once we
operationalise `word sense' as `dictionary word sense',
we now have a test that meets the challenge. The identity test for a
word sense in a particular dictionary is that two usages of the word belong to
it if and only if the lexicographer would have put them in the same
cluster.\footnote{A psycholinguistic investigation along these lines
is presented in \nocite{Jorgensen:90b} Jorgensen (1990).}

We can now present a different perspective on the ambiguity/generality
debate.  Where a word's uses fall into two entirely distinct clusters,
it is ambiguous, but where the clusters are less well-defined and
distinct, `vague' or `unspecified' may be a more appropriate
description. There is no reason to expect to find any clear
distinction between the two types of cases.

\section{Use, frequency, predictability, and the word sense}

`Clustering' is a metaphor.  It regards corpus lines as points in
space with measurable distances between them.  To give the account
substance, more must be said about the ways in which corpus lines may
be `close'.  In this section, I classify the types of relationships
that hold between a word's patterns of usage, and consider how these
considerations relate to lexicography.\footnote{I do not dwell on
cases of simple similarity, where there is a straightforward match
between corpus lines, or between a corpus line and a word's core
meaning.  While it is a major language-engineering problem to
operationalise even `simple similarity', it is not a problematic
matter, either theoretically or for lexicographers or other human
language users.}

There are five knowledge sources which come into play for 
understanding how a word contributes to the meaning or
communicative intent of the utterance or discourse it occurs in.  If a
word in context is interpretable by a language user, it will be by
virtue of these knowledge sources.

Whether a dictionary provides a word sense that matches an instance of use of the
word, is dictated by considerations of frequency and
predictability: if the instance exemplifies a pattern of use
which is sufficiently frequent, and is insufficiently predictable from
other meanings or uses of the word, then the pattern qualifies for treatment as
a dictionary sense.  A use is predictable, to the extent that a person
reading or hearing it for the first time can understand it (in all its
connotations).  Clearly, different dictionaries have different
thresholds of frequency and predictability. 

To illustrate the various processes whereby new types of usage may be
added to the repertoire for a word, let us consider the simple single-sense
word, {\em handbag}
	\begin{quote}	a small bag, used by women to carry money and personal
	things  (British; American English translation: purse)(LDOCE3) 
\end{quote}
As the 715 examples in the British National
Corpus (BNC)\footnote{For the BNC see {\tt http://info.ox.ac.uk/bnc}.
Counts were: {\em handbag} 609, {\em handbags} 103,
{\em handbagging} 1, {\em handbagged} 2.}  make plain, typical uses involve things being put into, or taken
out of, or looked for in handbags, or handbags being lost, found,
stolen, manufactured, admired, bought or sold.  But a couple of dozen examples
stretch the limits of the definition or fall outside it altogether.

First, a proper name, and a reference to a unique object:
\begin{verse}\small\sf 
the Drowning Handbag, an up-market eatery in the best part of town 
\\an inimitable rendering of the handbag speech in The
Importance of Being Earnest
\end{verse}
Next, metonymy, visual metaphor, simile:
\begin{verse}\small\sf 
She moved from handbags through gifts to the flower shop 
\\"How about you?   Did the bouncing handbag find  you?"\footnote{This
turns out to be a (sexist and homophobic) in-group joke, as well as a
case of both metonymy and of a distinct idiomatic use of the word.
Interestingly, in the text, ``the bouncing handbag'' succeeds in
referring, even though the idiom is not known to the addressee, as is
made explicit in the text.}
\\a weird, menacing building with bats hanging in the trees like handbags
\\Skin generally starting to age like old handbag or bodywork of
car 
\end{verse}
Next,  Mrs Thatcher:
\begin{verse}\small\sf 
from Edward Heath's hip-pocket to Margaret Thatcher's handbag
and on to Mr Major's glass of warm beer
\\``Thousands \ldots will be  disgusted at the way she [Thatcher]
is lining  her handbag''
\\send out Mrs Thatcher with a fully-loaded handbag
\\``If you want to define the Thatcher-and-after era in a single
 phrase'', he muses, `` `accountants with plenary
 powers' says it.'' Well now-- I would have gone for something a
little snappier: `A mad cow with a handbag,' comes to mind as a first attempt.
\\She [Thatcher] cannot see an institution  without hitting it with
her  handbag.

\end{verse}

The last of these is cited in another citation as the launching-point
of verbal {\em handbag}. Of the three verbal citations, all were
species of hitting and in two of them, Mrs. Thatcher was the
perpetrator.  

Next, and closely related to Mrs. Thatcher, `handbag
as weapon':

\begin{verse}\small\sf 
Meg swung her handbag.  
\\determined women armed with heavy handbags
\\it was time to race the old  ladies back to the village for the
 tea and scones of Beck Hall.   I beat them, but only just-- those
handbags are lethal. 
\\old ladies continue to brandish their handbags and umbrellas at the
likes of Giant Haystacks
\\ the blue rinse  brigade \ldots will be able to  turn out in force without having to  travel and give poor Louis
 Gerstner the handbagging of his  life. 
\\Peterborough manager Chris Turner added: "Evidently one of
their players  caught one of our players and 
 it was handbags at 10 paces and then someone threw a punch."
\end{verse}
The final, quite distinct group relates to discos, and the lexical
unit {\em dance round your handbag}, a pejorative phrase
for the behaviour of certain exclusively female groups at discotheques and
dances where ---prototypically--- they dance in a circle with their handbags on the floor in
the middle.  The conversational
speech subcorpus of the BNC provides two instances of the full form
while in the written corpus, the two related corpus lines, both from music
journalism, make only fleeting references to the collocation, and strikingly
indicate a process of lexicalisation:
\begin{verse}\small\sf 
The shoot was supposed to be a secret, but word got out
 and Hitman regulars travelled down to Manchester. Two thousand
couldn't  get into the club, and tension 
 mounted between trendy regulars (locked out of their own club)  and
the Hitman's handbag brigade 
(shut out of their programme).
\\New Yawk drawling rap over Kraftwerk's `The Model' just does
not work, no way, 
 no how.  Handbag DJs will love it.
\end{verse}

All these uses can be traced back to the standard
sense:  the potential for using the word in the nonstandard way, is 
(in varying degrees) {\bf predictable} from 
\begin{itemize}
\item its standard meaning and use 
\item general linguistic knowledge (eg.\ of processes of metonymy, regular
polysemy, and ellipsis, etc., and, in this case, the relation between
words for goods and words for shops or departments of shops where
those goods are sold),
\item general world knowledge (eg.\ regarding Mrs.\ Thatcher, or
juvenile female behaviour at discotheques) and
\item knowledge of related collocations (eg.\ ``lining their
pockets'', ``{\sc weapon} at {\sc number} paces'')
\item taxonomic knowledge 
\end{itemize}
These five knowledge sources define the conceptual space within which
lexical creativity and productivity, and the idea of a `word sense',
are located.\footnote{In \nocite{thesis} Kilgarriff (1992),
in the context of an analysis of polysemy, I call the first four knowledge
types {\sc homonymy, alternation, analogy} and {\sc
collocation}. (Taxonomy is addressed separately.)}

Needless to say, they frequently interact in complex ways.  In
``handbags at ten paces'', the speaker\footnote{This is presented as a
quotation of a football manager's spoken comment; quite whether it is
verbatim, or the Daily Telegraph journalist's paraphrase, we shall
never know.}  assumes the addressee's awareness of handbag-as-weapon.
Note that ``*briefcases at ten paces'' and ``*shoulder-bags at ten
paces'' do not carry the same meaning.  Although briefcases and
shoulder-bags are just as viable weapons as handbags, the words {\em
briefcase} and {\em shoulder-bag} do not carry the `weapon'
connotations which make the citation immediately understandable.
Handbag-as-weapon is a feature of the word, over and above the extent
to which it is a feature of the denotation.

In the citation's context, there is no overt reason for a reference to
{\em handbag}; the people involved are men, not women, so not
prototypical handbag-users, and there is no other reference to
femininity.  It would appear that the speaker is aiming
to both distance himself from and minimise the significance of the
incident by treating it as a joke.  The `duel' metaphor is itself a
joke, and the oddity of {\em handbag} in the context of either
football or duel, along with its associations with femininity and
Mrs. Thatcher, contributes to the effect.  Moreover, there is
a sexist implication that the men were behaving like women and thereby
the matter is laughable.  

Interpreting ``handbags at ten paces'' requires lexical knowledge of
``handbag-as-weapon'', collocational knowledge of both form and
meaning of ``{\sc weapon} at {\sc number} paces'', and (arguably)
knowledge of the association between handbags and models of
masculinity and femininity.

The `music journalism' use displays some further features.  {\em
Handbag} was lexicalised in the clubbing world in ca. 1990 as a music
genre: the genre that, in the 1970s and 1980s, certain classes of
young women would have danced round their handbags to.\footnote{Thanks
to Simon Shurville for sharing his expertise.} 
The coinage
emanates from the gay and transvestite club scene and is 
redolent with implications, from the appropriation of the handbag as a
symbol of gay pride, to changes in the social situation of women over
the last twenty years (and its expression in fashion accessories), to
transvestite fantasies of being naive seventeen-year-old girls in a
more innocent age.  

To restrain ourselves to more narrowly linguistic matters: the license
for the coinage is via the  ``dance round your handbag''
collocation, not directly from handbags.  As shown by the spoken
corpus evidence, the regular, non-ironic use of the collocation
co-exists with the music-genre use.  It is of much wider currency:
all but two of a range of informants knew the collocation, whereas
only two had any recollection of
the music-genre use.  Also, `handbag' music (or at least the use of
that label) was a 1990-91 fashion, and the term is no longer current:
1996 uses of it will probably refer back to 1990-91 (as well as back
to the 1970s and 1980s).

Syntactically, the most information-rich word of the collocation has
been used as a nominal premodifier for other nouns: in the music-genre
sense, it is used as other music-genre words, as an uncountable
singular noun, usually premodifying but potentially occurring on its
own: ``Do you like jazz/house/handbag?''

\subsection{Frequency}

These arguments make clear that there is a {\em prima facie} case for
including handbag-as-weapon and handbag-as-music-genre as dictionaries
senses, and ``dance round your handbag'' as an only partially
compositional collocation.  Each exhibits lexical meaning which is not
predictable from the base sense.  So why do the dictionaries not list
them?  The short answer is frequency.  Around 97\% of {\em handbag}
citations in the BNC are straightforward base sense uses. The
music-genre sense is certainly rare, possibly already obsolete, and
confined to a subculture.  The collocation is partially compositional
and occurs just twice in the corpus: for any single-volume
dictionary, there will not be space for vast numbers of partially
compositional collocations.  Not only is a lexicographer ``a
lexicologist with a deadline''
\cite{Fillmoretalk:88} but also a lexicologist with a page 
limit.\footnote{It is an interesting question, touched on in
\nocite{ak:chum} Kilgarriff (1993) but worthy of a much fuller investigation, what
the percentage of `anomalous' uses might be for various classes of
words.  One would expect the figures to be highly corpus-dependent.
A large proportion of the BNC is material written by novelists and journalists
---who earn their living, in some measure, through their skills in the
original and engaging use of language.  (The music-genre use of {\em
handbag} probably first occurred in advertising material, probably
the most fecund discourse of all.) Also one might expect spoken
material to have a higher proportion of set phrases, owing to the time
constraints on the production of spoken language.}

\subsection{Analytic definitions and entailments}

The handbag-as-weapon sense is rather more common, and a further
consideration comes into play.  The denotations of base-sense {\em
handbag} and handbag-as-(potential)-weapon are the
same. Correspondingly, the lexical fact that there is a use of {\em
handbag} in which it is conceptualised as a weapon does not render the
LDOCE definition untrue.  A lexicographer operating according to the
classical approach whose goal was simply to provide
necessary and sufficient conditions for identifying each word's
denotation would say that the `weapon' aspect of meaning was
irrelevant to his or her task.  A more pragmatic lexicographer
might also follow this line, particularly since space is always at a
premium.

The situation is a variant on autohyponymy \cite[pp
63--65]{Cruse:86}, the phenomenon of one sense being the
genus of another sense of the same word.  The prototypical example
is {\em dog} (canine {\em vs.}\ male canine).  {\em Dog} is a case
where there clearly are distinct senses.  For {\em knife} (weapon $vs.$
cutlery $vs.$ bladed object) \nocite{Cruse:95} Cruse (1995, pp 39--40) argues for
``an intermediate status'' between monosemy and polysemy, since, on
the  one hand, `bladed-object' is a coherent category which
covers the denotation, but on the other, in a scenario where there was
a penknife but no cutlery knife at a table setting, one might
reasonably say ``I haven't got a knife''. COBUILD2
distinguishes `weapon' and `cutlery' senses, while LDOCE3 provides a
single, analytically adequate, `bladed object' sense. 

In a discussion of the polysemy of {\em sanction},
\nocite{Kjellmer:93} Kjellmer (1993) makes a related observation.  His goal is to
examine how language breakdown is avoided when a word has antagonistic
readings.  Nominal {\em sanction} is such a word: in ``sanctions
imposed on Iraq'' the meaning is akin to punishment (`{\sc pun}')
whereas in ``the proposal was given official sanction'' it is related
to endorsement (`{\sc end}').  A first response is that the context
disambiguates - punishment, not support, is the sort of thing you
``impose'', whereas ``give'' implies, by default, a
positively-evaluated thing given.  Syntax is also a clue: the plural
use is always {\sc pun}, whereas determinerless singular uses suggest
{\sc end}.  Kjellmer then finds the following instances:
\begin{verse}\small\sf
The process of social control is operative insofar as sanction plays a
part in the individual's behaviour, as well as in the group's behaviour.
By means of this social control, deviance is either eliminated or
somehow made compatible with the function of the social group.\\
Historically, religion has also functioned as a tremendous engine of
vindication, enforcement, sanction, and perpetuation of various other
institutions.
\end{verse}
Here the context does not particularly favour either reading against
the other. In the second case, the co-ordination with both an {\sc
end} word ({\em vindication}) and a {\sc pun} one ({\em enforcement})
supports both readings simultaneously.  How is this possible, given
their antagonism? How come these uses do not result in ambiguity and
the potential for misinterpretation?  The answer seems to be that,
\begin{quote}
 we may operate, as readers or listeners, at a general, abstract level
and take the word to mean `control, authority' until the context
specifies for us which type of control is intended, if indeed
specification is intended. In other words, faced with the dual
semantic potentiality of the word, we normally stay at a higher level
of abstraction, where the danger of ambiguity does not exist, until
clearly invited to step down into specificity. (p 120)\footnote{Kjellmer implies
that the further specification is a temporal process,
there being a time in the
interpretation process when the lexical meaning of the word is
accessed but specified for `control' but not for either {\sc pun} or {\sc end}.  I see no
grounds for inferring the temporal process from the logical structure.} 
\end{quote}
Citations where {\em sanction} is unspecified for either {\sc pun} or
{\sc end} are rare, and there is no case for including the
unspecified `control' sense in a dictionary.

The example demonstrates a relationship between a lexicographer's
analytic defining strategy and the interpretation process.  There are
occasions where a `lowest common denominator' of the usually distinct standard uses of
a word will be the appropriate reading, in a process analogous to the
way an analytically-inclined lexicographer might write a definition
for a word like {\em charming} or {\em knife}, which would cover the
word's uses in two or more
distinct corpus clusters.  Some dictionaries use nested entries as a
means of representing meanings related in this way.

\section{Implications for WSD}

The argument so far exposes a lack of foundations to the
concept of `word sense'.   But, a WSD researcher might say, ``so
what?''  What are the implications for practical work in disambiguation?

The primary implication is that a task-independent set of word senses
for a language is not a coherent concept.  Word senses are simply
undefined unless there is some underlying rationale for clustering,
some context which classifies some distinctions as worth making and
others as not worth making.  For people, homonyms like {\em pike} are
a limiting case: in almost any situation where a person considers it
worth their while attending to a
sentence containing {\em pike}, it is also worth their while making the
fish/weapon distinction. 

Lexicographers are aware of this: the senses they list are
selected according to the editorial policy and anticipated users and
uses of the particular dictionary they are writing.  Until recently,
WSD researchers have generally proceeded as if
this was not the case: as if a single program ---disambiguating,
perhaps, in its English-language version, between the senses given in
some hybrid descendant of Merriam-Webster, LDOCE, COMLEX, Roget,
OALDCE and WordNet ---would be relevant to a wide range of NLP
applications.


There is no reason to expect the same
set of word senses to be relevant for different tasks.

The handbag data shows how various the non-standard uses of {\em
handbag} are.  These uses are sufficiently predictable or
insufficiently frequent to be dictionary senses (in a dictionary such
as LDOCE).  They are licensed by a combination of linguistic
principles, knowledge of collocations and lexico-syntactic contexts,
and world knowledge.  Only in a single case, the department store
metonym, is there a plausible linguistic principle for extending the
base meaning to render the non-standard use interpretable.  The data
suggest that little coverage will be gained by an NLP system
exploiting generative principles which dictate meaning potential.
The non-standard uses of words tend to have their own
particular history, with one non-standard use often built on another,
the connections being highly specific to a word or lexical field.

The handbag data also indicates how the corpus dictates the word
senses.  The BNC is designed to cover a wide range of standard
English, so is consonant with a general purpose dictionary.  The
common uses in the one should be the senses in the other.  But, were
we to move to a music journalism corpus, the music-genre sense would
be prominent.  A 1990s music-journalism dictionary would include it.


The practical method to extend the coverage of NLP systems to
non-standard uses is not to compute new meanings, but to list them.
Verbal {\em handbag} can, if sufficiently frequent, be added to the
lexicon as a synonym for {\em beat}; ``{\sc weapon} at {\sc number}
paces'' as one for ``have an argument''.  For the medium term future,
the appropriate language-engineering response to a use of a word or
phrase which the system needs to interpret but which it is currently
misinterpreting because the word or phrase's use does not match that
in the lexicon, is to add another lexical entry.\footnote{A
well-organised, hierarchical lexicon will mean that this need not
introduce redundancy into the lexicon.}

The implications of the account for different
varieties of NLP application are addressed in Kilgarriff (1997a, 1997b).
\nocite{thai,frascati}

\section{Conclusion}

Following a description of the conflict between WSD and lexicological
research, I examined the concept, `word sense'.  It was not found to
be sufficiently well-defined to be a workable basic unit of meaning.

I then presented an account of word meaning in which `word sense' or
`lexical unit' is not a basic unit.  Rather, the basic units are
occurrences of the word in context (operationalised as corpus
citations).  In the simplest case, corpus citations fall into one or
more distinct clusters and each of these clusters, if large enough and
distinct enough from other clusters, forms a distinct word sense.  But
many or most cases are not simple, and even for an apparently
straightforward common noun with physical objects as denotation, {\em
handbag}, there are a significant number of aberrant citations.  The
interactions between a word's uses and its senses were explored in some
detail.  The analysis also charted the potential for lexical
creativity.

The implication for WSD is that word senses are only
ever defined relative to a set of interests.  The set of senses
defined by a dictionary may or may not match the set that is relevant
for an NLP application.  

The scientific study of language should not include word senses as
objects in its ontology.  Where `word senses' have a role to play in a
scientific vocabulary, they are to be construed as abstractions over
clusters of word usages. The non-technical term for ontological commitment is
`belief in', as in ``I (don't) believe in ghosts/God/antimatter''.
One leading lexicographer doesn't believe in word senses.
I don't believe in word senses, either.

\subsection*{Acknowledgments}
This research was supported by the {\sc epsrc} Grant \mbox{K18931}, {\em
SEAL.}  I would also like
to thank Sue Atkins, Roger Evans, Christiane Fellbaum, Gerald Gazdar,
Bob Krovetz, Michael Rundell, Yorick Wilks and the anonymous reviewers
for their valuable comments.

\bibliographystyle{fullname}

\end{document}